\documentclass[10pt]{article}

\usepackage[a4paper,margin=0.75in]{geometry}
\usepackage{graphicx}
\usepackage{amsmath}
\usepackage{amssymb}
\usepackage{booktabs}
\usepackage{multirow}
\usepackage{float}
\usepackage{microtype}
\usepackage{balance}
\usepackage{url}
\usepackage{dblfloatfix}
\usepackage{caption}
\usepackage{titlesec}
\usepackage{authblk}
\pdfoutput=1

\titleformat{\subsection}
{\small\bfseries}
{\thesubsection}
{0.5em}
{}

\titlespacing*{\subsection}
{0pt}
{0.5em}
{0.3em}

\title{YOLO-AMC: An Improved YOLO Architecture with Attention Mechanisms for Building Crack Detection}

\author[1]{Ching-Yu Tsai}
\author[1]{Chia-Min Lin}
\author[1]{Chih-Hsiang Yang}
\author[1]{Yung-Che Wang}
\author[1]{Jen-Shiun Chiang}

\affil[1]{Department of Electrical and Computer Engineering, Tamkang University, Taiwan}

\date{}

\begin{document}

\twocolumn[
\maketitle

\begin{abstract}
Crack detection plays an important role in infrastructure inspection and Structural Health Monitoring (SHM). However, traditional manual inspection methods are time-consuming and labor-intensive, making them difficult to scale to large-area monitoring scenarios. In addition, cracks typically appear as thin, low-contrast structures and are easily affected by background noise, which poses challenges for existing object detection models in terms of fine feature representation and precise localization. This study proposes an improved YOLO-based architecture with 1integrated attention mechanisms, termed YOLO-AMC (YOLO with Attention Mechanisms for Crack Detection), to enhance automated crack detection performance. Based on YOLOv11, the original C2PSA module is removed, and multiple attention mechanisms are introduced into the multi-scale feature fusion layers of the Neck, including Global Attention Mechanism (GAM), Residual Convolutional Block Attention Module (Res-CBAM), and Shuffle Attention (SA), to strengthen cross-scale feature integration. Experimental results demonstrate that the proposed YOLO-AMC architecture consistently outperforms the baseline YOLOv11n and YOLOv8n models across multiple evaluation metrics. Among the evaluated attention modules, GAM achieves the best detection performance, obtaining mAP@0.5 = 0.9917 and mAP@0.5:0.95 = 0.9506 on the test dataset, which are higher than those of YOLOv11n (0.9833 / 0.9112) and YOLOv8n (0.9707 / 0.8921). Furthermore, while maintaining a reasonable computational complexity of 7.6 GFLOPs, the proposed model achieves practical inference performanceon both high-performance GPU platforms and resource-constrained edge devices, reaching 110.95 FPS on an NVIDIA RTX 4090 platform and approximately 5 FPS on a Raspberry Pi 5 edge device. These results demonstrate a favorable trade-off between accuracy and deployment efficiency. The implementation code for this study is available on GitHub at \url{https://github.com/CY-Tsai24/YOLO-AMC}.
\end{abstract}

\vspace{0.5em}

\vspace{1em}
]

\begin{figure*}[t]
\centering
\includegraphics[width=\textwidth]{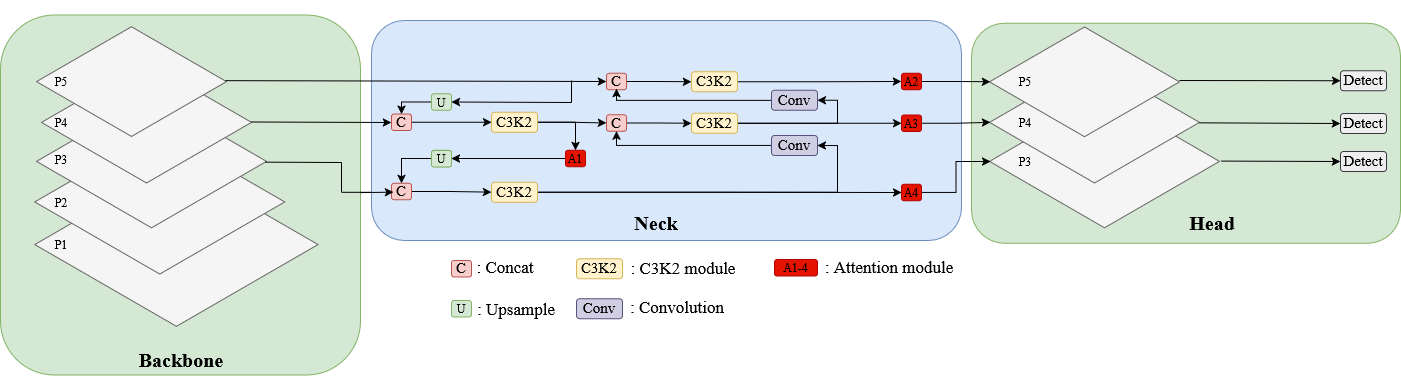}
\caption{Overall architecture of the proposed YOLO-AMC framework and the attention insertion positions in the Neck structure.}
\label{fig1}
\end{figure*}

\section{Introduction}
\label{sec:introduction}

Crack detection plays an important role in infrastructure inspection and Structural Health Monitoring (SHM). Surface cracks are widely recognized as early indicators of material degradation and potential structural failure, making them critical for safety evaluation and maintenance decision-making in bridges, roads, and buildings. Traditional crack inspection relies mainly on manual visual inspection, which is time-consuming, labor-intensive, and susceptible to subjective judgment and fatigue, making it difficult to apply to large-scale infrastructure monitoring.

With advances in computer vision and deep learning, automated crack detection methods based on convolutional neural networks (CNNs) have increasingly been adopted in place of traditional image processing approaches. Cha \textit{et al.} \cite{cha2018autonomous} employed Faster R-CNN for structural damage detection and demonstrated the effectiveness of deep learning in identifying multiple types of defects. Fan \textit{et al.} \cite{fan2019road} utilized deep convolutional neural networks for road crack recognition, showing that deep features can significantly improve detection accuracy. As real-time inspection systems become increasingly important in practical applications, achieving high detection accuracy while maintaining efficient inference speed has become a key research issue \cite{peng2024smart}.

Among existing object detection models, the YOLO (You Only Look Once) series \cite{mohammed2025architecture},\cite{ali2024yolo} has been widely adopted due to its favorable balance between accuracy and speed. However, in crack detection scenarios, cracks usually appear as thin, low-contrast, and irregular structures, which makes feature extraction and localization more challenging for standard YOLO architectures. YOLOv11 \cite{ultralytics2025} introduces the C2PSA module to enhance spatial awareness. However, fine crack features may still be weakened during multi-scale downsampling, leaving room for further improvement in balancing detection accuracy and computational cost.

To enhance the representational capability of convolutional neural networks for critical features, attention mechanisms have been widely integrated into object detection architectures. By reweighting feature responses in channel and spatial dimensions, attention modules enable the network to focus on informative regions while suppressing background noise. Previous studies have shown that embedding attention modules into YOLO-based models can improve the detection of small-scale or weak-feature targets \cite{cai2024lightweight,zamri2024enhanced,rupak2024investigation}. In our previous work \cite{lai2025yolo}, integrating attention modules into the YOLOv11 architecture improved crack detection performance while maintaining practical inference efficiency. These findings indicate that appropriate attention design can further enhance detection accuracy without significantly increasing model complexity.

Although attention mechanisms have shown promising results, designing an efficient attention-enhanced YOLO architecture that improves detection performance while maintaining computational efficiency remains an open problem. Therefore, this study proposes an improved YOLO-based architecture with integrated attention mechanisms, termed YOLO-AMC (YOLO with Attention Mechanisms for Crack Detection), built upon the YOLOv11 framework. Figure~\ref{fig1} illustrates the overall architecture of the proposed framework, in which attention modules are incorporated into the Neck to enhance the fusion of crack-related features across multiple scales.

The main contributions of this work are summarized as follows:

\begin{enumerate}
    \item A modified YOLO architecture named YOLO-AMC is proposed by integrating multiple attention mechanisms into YOLOv11 for crack detection, achieving a balance between detection accuracy and inference efficiency.
    
    \item The experimental results demonstrate that the proposed model consistently outperforms the baseline models of YOLOv11 and YOLOv8 in crack detection performance.
    
    \item The proposed architecture is deployed and evaluated on edge devices such as Raspberry Pi 5, demonstrating stable performance under limited computational resources and strong potential for practical SHM edge applications.
\end{enumerate}

\newpage
\section{Related Works}
\label{sec:relatedworks}

\subsection{Deep Learning for Crack Detection}

As infrastructure continues to age, crack detection has become an important research topic in the field of structural health monitoring. Early crack detection methods mainly relied on traditional image processing techniques and handcrafted features. However, under complex backgrounds, illumination variations, or low-contrast conditions, these approaches often fail to maintain stable detection performance. In recent years, with advances in deep learning, convolutional neural networks (CNNs) have been widely applied to crack detection tasks and have demonstrated significant improvements in both accuracy and generalization performance \cite{cha2017deep,fan2018automatic}.

To improve crack localization capability, subsequent studies have gradually extended from image classification to object detection and pixel-level segmentation approaches. Li \textit{et al.} \cite{li2019automatic} employed a Fully Convolutional Network (FCN) to localize damage regions, while Yang \textit{et al.} \cite{yang2019feature} utilized feature pyramid structures and hierarchical feature fusion to enhance the representation of multi-scale cracks. Their results showed that multi-level feature integration plays a critical role in detecting fine crack patterns. Although pixel-level segmentation models such as U-Net can provide more precise contour information, they usually involve higher computational cost, which limits their applicability in real-time detection or edge-device deployment scenarios \cite{li2018pyramid}.

\subsection{YOLO-Based Object Detection for Crack Detection}

In the field of object detection, the YOLO (You Only Look Once) series has been widely adopted for real-time vision tasks due to its one-stage detection architecture, which provides a favorable balance between inference speed and detection accuracy. Since the introduction of multi-scale prediction in YOLOv3, small-object detection performance has been significantly improved \cite{redmon2018yolov3}. YOLOv4 further enhanced performance through network architecture optimization and improved training strategies, resulting in better stability and accuracy in practical applications \cite{bochkovskiy2020yolov4}. Zhou \textit{et al.} \cite{zhou2023automatic} applied YOLOv4 to tunnel crack detection and demonstrated the feasibility of the YOLO framework for engineering inspection tasks.

Although multi-scale prediction improves the detection of small objects, cracks typically exhibit thin, low-contrast, and irregular patterns. After multiple stages of downsampling and feature fusion, weak features are easily suppressed by high-level semantic information and background noise \cite{li2026lightweight,liang2022edge}. In most YOLO-based architectures, the Neck is responsible for cross-scale feature fusion, typically implemented using feature pyramid structures and a Path Aggregation Network (PANet) to integrate multi-level information. Features at different levels contain distinct spatial resolutions and semantic representations, and if enhancement is not applied at appropriate stages, fine crack features may be weakened during the fusion process, thereby affecting overall detection performance.

\subsection{Attention Mechanisms in Object Detection Networks}

To enhance the representational capability of deep neural networks for critical features, attention mechanisms have been widely introduced into various vision models in recent years. CBAM (Convolutional Block Attention Module) \cite{woo2018cbam} improves feature discrimination by sequentially applying channel attention and spatial attention, enabling the network to focus on more informative regions. GAM (Global Attention Mechanism) \cite{liu2021global} emphasizes global interactions between channel and spatial information to strengthen cross-dimensional feature relationships. Shuffle Attention (SA) \cite{zhang2021sanet} adopts channel grouping and feature reorganization to enhance feature representation while maintaining a lightweight architecture.

Attention modules usually adjust feature distributions through reweighting operations, and their effectiveness is closely related to the original network structure. Previous studies have shown that in networks with residual connections, the modulation effect of attention modules may be influenced by the main branch signal \cite{wang2017residual,hu2018squeeze}, indicating that the performance gains of attention mechanisms depends not only on the module design itself but also on the integration strategy and insertion level within the network. In object detection tasks, attention mechanisms have been integrated into YOLO-based models to improve the detection of small-scale or weak-feature targets. Chien \textit{et al.} \cite{chien2025yolov8am} embedded multiple attention modules into the YOLO architecture and achieved improved performance in detecting elongated objects. Similar approaches have also been applied to crack detection and pavement defect inspection tasks \cite{dong2025study,kim2021ecap}, demonstrating that attention mechanisms are effective for enhancing low-contrast and fine-structure features. With the increasing demand for edge-device deployment, recent studies have further focused on balancing model efficiency and feature enhancement under limited computational resources \cite{tang2024hic,wang2022trc}. Saeheaw \textit{et al.} \cite{saeheaw2025hfe} introduced feature enhancement and attention modules into the feature fusion layers of YOLOv11, showing that strengthening features at appropriate levels can improve detection performance while maintaining acceptable computational cost.

In addition, multi-scale feature fusion plays a critical role in small-object detection. Features at different levels contain different spatial resolutions and semantic information, and without appropriate enhancement at fusion nodes, weak features may be suppressed during the aggregation process \cite{li2018pyramid,li2019selective}. Lin \textit{et al.} \cite{lin2017fpn} and Tan \textit{et al.} \cite{tan2020efficientdet} pointed out that the design of feature pyramid structures and module configurations directly affects both detection accuracy and inference efficiency. Therefore, when designing attention-enhanced detection architectures, it is necessary to consider the balance among module type, insertion position, and computational cost.

\begin{figure*}[!t]
\centering
\includegraphics[width=\textwidth]{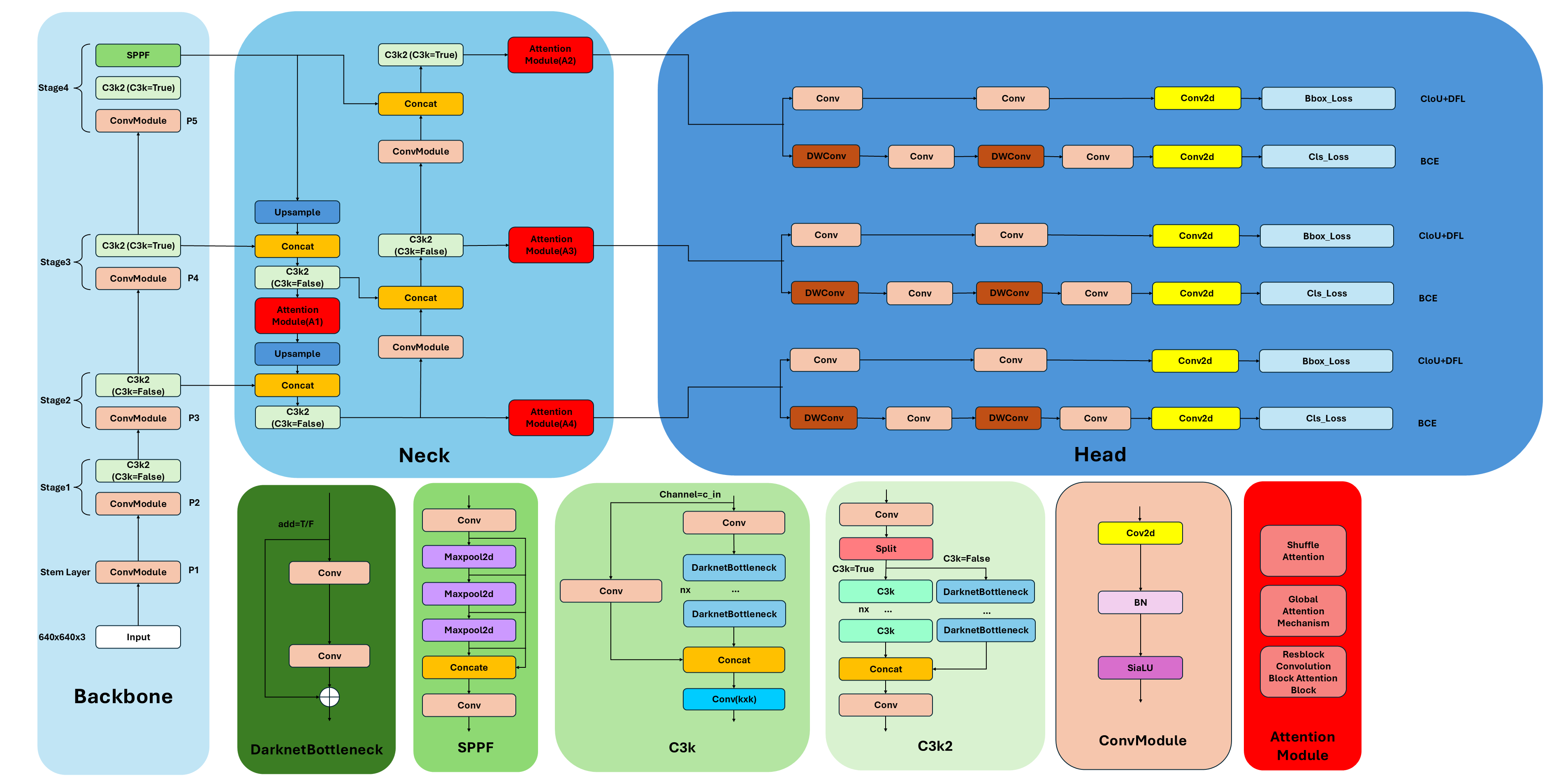}
\caption{Architecture of the proposed YOLO-AMC model. Attention modules are inserted at different positions (A1--A4) within the Neck to enhance multi-scale feature fusion.}
\label{fig2}
\end{figure*}

\section{Methodology}
\label{sec:methodology}

This study aims to improve the performance of YOLO-based architectures for crack detection tasks and proposes an enhanced model, named YOLO-AMC, which integrates attention mechanisms into the baseline detection framework. Built upon the YOLOv11 framework, this work incorporates Global Attention Mechanism (GAM), Shuffle Attention (SA), and Residual-CBAM (Res-CBAM) into the network and systematically investigates the influence of different attention integration strategies on detection accuracy and inference efficiency.

\subsection{YOLOv11 Baseline Architecture}

In this study, YOLOv11 is adopted as the baseline architecture. The model consists of three main components, namely the Backbone, Neck, and Head, which are responsible for hierarchical feature extraction, multi-scale feature fusion, and final classification and bounding box regression, respectively. Following the design philosophy of the YOLO series for real-time object detection, YOLOv11 provides a favorable balance between model complexity and inference efficiency, making it suitable for real-time inspection scenarios.

The original YOLOv11 architecture introduces the C2PSA (Cross-Stage Partial with Spatial Attention) module in the Backbone to enhance spatial feature representation. However, in crack detection tasks, cracks usually appear as thin and low-contrast structures, whose discriminative information relies heavily on fine textures and weak feature responses. Under such conditions, the built-in attention mechanism may interact with externally introduced attention modules, making it difficult to independently evaluate the contribution of different attention designs to feature enhancement.

To ensure a fair comparison, the C2PSA module in the original YOLOv11 architecture is removed in this study, resulting in a simplified baseline model without built-in attention structures. This modification allows the subsequently introduced attention modules to be evaluated under consistent architectural conditions for systematic comparison. For overall performance evaluation, the original YOLOv11 and YOLOv8 are used as the primary baseline models, and their performance is compared with that of the proposed YOLO-AMC architecture to verify the effectiveness of the improved model in crack detection tasks as well as its feasibility for practical deployment.

\subsection{Attention Modules}

To analyze the behavioral differences of various attention mechanisms within the YOLO architecture, three representative attention modules with distinct structural characteristics were selected: the Global Attention Mechanism (GAM) \cite{liu2021global}, Shuffle Attention (SA) \cite{zhang2021sanet}, and Residual-CBAM (Res-CBAM) \cite{chien2025yolov8am}. Res-CBAM is derived from the original CBAM \cite{woo2018cbam} and is implemented following the design adopted in YOLOv8-AM.

Let the input feature map be defined as:

\begin{equation}
F \in \mathbb{R}^{C \times H \times W}
\label{eq1}
\end{equation}

\noindent where $C$, $H$, and $W$ denote the number of channels, height, and width of the feature map, respectively. The objective of the attention module is to reweight the feature map $F$ to enhance discriminative channels or spatial regions, thereby enhancing the representational capability of the network.

\subsubsection{Global Attention Mechanism (GAM)}

The Global Attention Mechanism (GAM) models global dependencies across both channel and spatial dimensions. It performs feature modulation by sequentially applying channel attention and spatial attention, which can be expressed as follows.

\begin{equation}
F_c = A_c(F) \otimes F
\label{eq2}
\end{equation}

\begin{equation}
F_{out} = A_s(F_c) \otimes F_c
\label{eq3}
\end{equation}

\noindent where $A_c(\cdot)$ and $A_s(\cdot)$ denote the channel attention and spatial attention functions, respectively, and $\otimes$ represents element-wise multiplication. Since GAM models feature relationships through global feature interactions, the resulting attention weights depend on the overall feature distribution. Therefore, different feature levels may exhibit different enhancement effects, and the global modeling process may introduce additional computational cost.

\subsubsection{Shuffle Attention (SA)}

Shuffle Attention (SA) adopts a group-wise strategy combined with channel splitting to perform attention modeling with reduced computational complexity. The input feature map is divided into multiple channel groups, and both channel attention and spatial attention are applied within each group. This design limits the scope of attention computation while providing effective feature modulation, allowing the module to maintain low computational cost while enhancing feature representation \cite{zhang2021sanet}.

Let the input feature map $F$ be divided into $G$ groups along the channel dimension:

\begin{equation}
F = [F_1, F_2, \ldots, F_G]
\label{eq4}
\end{equation}

For the $g$-th feature group, a channel split operation is applied to further divide it into two branches:

\begin{equation}
F_{g1}, F_{g2} = \mathrm{Split}(F_g)
\label{eq5}
\end{equation}

\noindent where $F_{g1}$ and $F_{g2}$ are used for channel attention and spatial attention modeling, respectively. The subsequent feature modulation process can be expressed as follows.

\begin{equation}
F_{g1}^{'} = A_c(F_{g1}) \cdot F_{g1}, \quad
F_{g2}^{'} = A_s(F_{g2}) \cdot F_{g2}
\label{eq6}
\end{equation}

The final output feature is obtained through channel recombination using the channel shuffle operation:

\begin{equation}
F_{out} = \mathrm{ChannelShuffle}(\mathrm{Concat}([F_{g1}^{'}, F_{g2}^{'}]))
\label{eq7}
\end{equation}

SA adopts a group-wise and localized design, limiting the scope of attention computation to reduce computational cost. This lightweight architecture enables effective feature enhancement while maintaining computational efficiency, making it suitable for deployment in resource-constrained edge-device scenarios.

\subsubsection{Residual Block Attention Module (ResBlock-CBAM)}

ResBlock-CBAM is a residual attention module derived from the original Convolutional Block Attention Module (CBAM) \cite{woo2018cbam}. In this study, the implementation follows the Res-CBAM design adopted in YOLOv8-AM \cite{chien2025yolov8am}, where channel attention and spatial attention are integrated with a residual connection to preserve the direct propagation of the original feature information.

The operation of ResBlock-CBAM can be expressed as follows.

\begin{equation}
F^{'} = A_c(F) \otimes F
\label{eq10}
\end{equation}

\begin{equation}
F^{''} = A_s(F^{'}) \otimes F^{'}
\label{eq11}
\end{equation}

\begin{equation}
F_{out} = F + F^{''}
\label{eq12}
\end{equation}

\noindent where $F^{'}$ denotes the intermediate feature refined by channel attention, and $F^{''}$ represents the feature further modulated by spatial attention.

By introducing a residual connection, the attention-refined features are added to the original input feature, helping preserve preserve low-level and mid-level feature information while stabilizing feature propagation throughout the network. Compared with attention modules without residual connections, ResBlock-CBAM allows the network to retain the original feature flow while applying attention modulation, which thereby reducing sensitivity to variations across feature levels. As a result, the module tends to provide more stable feature refinement across different feature scales and may improve the robustness of multi-scale feature fusion in object detection tasks.

\begin{table*}[!t]
\caption{Performance, model complexity, and inference speed comparison of YOLO-AMC variants and baseline models.}
\label{tab1}
\centering
\begin{tabular}{lccccccc}
\hline
Model & Parameters & FLOPs(G) & mAP@50 & mAP@50:95 & Precision & Recall & FPS \\
\hline
GAM & 3,021,587 & 7.6 & 0.9917 & 0.9506 & 0.9829 & 0.9866 & 110.95 \\
Res-CBAM & 3,573,211 & 8.7 & 0.9903 & 0.9430 & 0.9817 & 0.9842 & 108.32 \\
SA & 2,340,523 & 6.2 & 0.9893 & 0.9398 & 0.9835 & 0.9810 & 118.74 \\
YOLOv11 & 2,590,035 & 6.4 & 0.9833 & 0.9112 & 0.9717 & 0.9720 & 113.84 \\
YOLOv8 & 3,011,043 & 8.2 & 0.9707 & 0.8921 & 0.9738 & 0.9480 & 138.78 \\
\hline
\end{tabular}
\end{table*}

\subsection{Proposed Method}

In recent years, attention mechanisms have demonstrated strong performance in object detection tasks \cite{chien2025yolov8am,jiang2022improved,li2020adaptive,zhang2019featurefusion}. By incorporating attention mechanisms, models can more effectively enhance the representation of critical features while suppressing irrelevant or noisy information, thereby improving overall detection performance.

In this study, attention modules are integrated into the Neck structure of YOLOv11, and the original C2PSA module is removed to facilitate the evaluation of the contribution of external attention mechanisms to model performance. The Neck is selected as the insertion location because it serves as the core component responsible for multi-scale feature fusion in the YOLO architecture. Since cracks typically exhibit cross-scale distributions and subtle visual characteristics, their discriminative information may appear in feature maps of different resolutions. Therefore, introducing attention mechanisms during the feature fusion process may enhance the integration of crack-related features across multiple scales.

To construct an attention-enhanced architecture suitable for crack detection, four candidate insertion positions are defined in the Neck module of YOLOv11, denoted as A1, A2, A3, and A4, as illustrated in Figure~\ref{fig2}. These positions are located at different nodes along the multi-scale feature fusion path and correspond to feature maps with different spatial resolutions and semantic levels within the feature pyramid. In the YOLO-AMC architecture, feature maps at different levels exhibit significant variations in spatial resolution and semantic abstraction. Specifically, A1 corresponds to higher-resolution feature maps that preserve more local texture information, whereas A2 and A3 are located at intermediate levels that balance spatial detail and semantic information. In contrast, A4 corresponds to lower-resolution feature maps with stronger semantic representation. Following the multi-scale feature fusion process, the detection head utilizes feature maps from the feature pyramid, denoted as P3, P4, and P5, as shown in Figure~\ref{fig2}, to perform the final object detection. These feature levels integrate information from different stages of the network, and their effectiveness may be influenced by the configuration of attention modules at the A1--A4 positions. By inserting attention modules at different levels, multiple YOLO-AMC variants can be constructed to systematically evaluate the impact of different attention integration strategies on crack detection performance, as well as to analyze the interaction between attention mechanisms and feature hierarchy.

Under the same training conditions, YOLO-AMC variants are compared with the original YOLOv11 and YOLOv8 baseline models to evaluate the effectiveness of the proposed architecture in terms of detection accuracy and inference efficiency.

\section{Experiments}
\label{sec:experiments}

\subsection{Experimental Setup}

To systematically evaluate the performance of the proposed model for crack detection, all experiments were conducted using the same dataset, identical training settings, and the same model scale to ensure a fair and reproducible comparison.

\subsubsection{Dataset Description}

The composite crack dataset established in our previous study is used as the benchmark dataset in this work. The dataset is constructed by integrating five public crack image datasets, including BAC\_HIEN\_Crack\_Concrete\_2024 \cite{roboflow2024bac_hien}, Crack Detection.v2 \cite{roboflow2024crackv2}, Crack Detection.v3i \cite{roboflow2024crackv3i}, Crack Finder.v1i \cite{roboflow2024crackfinder}, and the Concrete Crack Images for Classification dataset \cite{ozgenel2025concrete}. The integrated dataset contains 15,524 training images, 2,276 validation images, and 1,665 testing images. All crack instances are manually re-annotated using bounding boxes to fit the object detection framework. To prevent data leakage between different source datasets, the integrated dataset was re-split after merging to ensure that each image appears in only one subset.

\subsubsection{Training Configuration}

All models were trained using the Ultralytics YOLOv11 framework. The experiments were conducted on a system equipped with an NVIDIA RTX 4090 GPU, an Intel Core i9-14900K CPU, and 64 GB of memory. The input image size was set to $640 \times 640$, and the batch size was set to 32. Each model was trained for up to 800 epochs, and an early stopping mechanism (patience = 100) was employed to prevent overfitting.

The optimizer and related hyperparameters followed the default Ultralytics settings, including an initial learning rate of 0.01, momentum of 0.937, and weight decay of 0.0005. To ensure reproducibility, the random seed was fixed at 0 for all experiments. All comparison models used the same training settings and dataset configuration to ensure that performance differences were primarily attributable to the model architecture design.

\subsubsection{Evaluation Metrics}

To comprehensively evaluate model performance, the following metrics are adopted:

\begin{itemize}
    \item mAP@0.5
    \item mAP@0.5:0.95
    \item Precision
    \item Recall
\end{itemize}

The inference time is defined as the total latency of the complete inference pipeline, including preprocessing, model forward pass, and post-processing (e.g., Non-Maximum Suppression). To ensure fair measurement, result saving is disabled during inference. All inference times are measured on a per-image basis using single-image inference mode. The average inference time and corresponding frames per second (FPS) are reported as efficiency metrics.

\subsection{Baseline Attention Validation}

After proposing the YOLO-AMC architecture, the attention-integrated YOLO variants are first compared with the original YOLOv11 and YOLOv8 under identical experimental settings in terms of detection accuracy, model complexity, and inference performance, to evaluate the effectiveness of the proposed YOLO-AMC for crack detection.

Table~\ref{tab1} summarizes the comparison results of YOLOv11, YOLOv8, and the YOLO-AMC variants with three attention modules (GAM, Res-CBAM, and SA) in terms of parameter count, FLOPs, detection accuracy, and inference speed. The results show that all attention-integrated models outperform the original YOLOv11 and YOLOv8 in both mAP@0.5 and mAP@0.5:0.95. Among them, GAM achieves the highest detection accuracy (mAP@0.5 = 0.9917, mAP@0.5:0.95 = 0.9506), followed by Res-CBAM and SA. These results indicate that attention mechanisms can effectively enhance the representation of fine crack features.

In terms of model complexity, the number of parameters and FLOPs increases after the integration of attention modules; however, all models remain within the lightweight model range. Notably, SA achieves higher detection accuracy while requiring fewer parameters and FLOPs than YOLOv11, suggesting that the performance improvement is not solely attributable to increased model scale but is also related to the feature reweighting capability of the attention mechanism. Although GAM and Res-CBAM introduce additional computational overhead, their overall model complexity remains within an acceptable range.

For inference efficiency, end-to-end inference time is measured on the NVIDIA RTX 4090 GPU using the full test set (1665 images). The definition of inference time follows Section~4.1.3. The results show that although the FPS slightly decreases after adding attention modules compared with YOLOv11 and YOLOv8, all YOLO-AMC variants still maintain efficient inference performance. Although YOLOv8 achieves the highest inference speed, SA ranks second (118.74 FPS) and provides a better balance between detection performance and computational cost.

\subsection{Ablation Study}

This section analyzes the effect of attention design in YOLO-AMC through ablation experiments, including attention placement, multi-layer configuration, and hybrid attention integration.

\subsubsection{Attention Placement Analysis}

Three attention modules (GAM, Res-CBAM, and SA) are inserted into four feature levels (A1--A4) in the Neck, and their detection performance is compared.

According to Table~\ref{tab2}, the GAM module achieves its best performance is achieved at position A3 (0.9441), while A2 yields the lowest result (0.9408), with a maximum difference of 0.0033. This indicates that GAM is more sensitive to changes in feature levels. The result suggests that global attention modeling does not produce consistent feature enhancement across different semantic levels. In contrast, Res-CBAM exhibits very small performance variations across the four insertion positions, with a maximum difference of only 0.0008. This indicates that Res-CBAM has substantially lower dependence on insertion location. Such stability may help maintain good detection performance under different structural configurations. SA exhibits an intermediate behavior between GAM and Res-CBAM, with a maximum difference of 0.0029, indicating a moderate sensitivity to feature-level changes.

It is worth noting that position A3 achieves relatively better performance for all three attention modules. This suggests that the intermediate feature level may provide a better balance between spatial detail and high-level semantic information, allowing the attention mechanism to enhance crack-related features more effectively. This observation supports the importance of considering feature hierarchy when selecting attention insertion positions in the YOLO-AMC architecture.

To objectively analyze the influence of different insertion positions on the performance variation of YOLO-AMC, two placement sensitivity metrics are adopted in this study.

\begin{table}[!ht]
\caption{Comparison of mAP@0.5:0.95 for different attention modules inserted at single-layer positions (A1--A4).}
\label{tab2}
\centering
\begin{tabular}{lcccc}
\hline
Attention & A1 & A2 & A3 & A4 \\
\hline
GAM & 0.9429 & 0.9408 & 0.9441 & 0.9421 \\
Res-CBAM & 0.9414 & 0.9412 & 0.9420 & 0.9419 \\
SA & 0.9392 & 0.9389 & 0.9418 & 0.9391 \\
\hline
\end{tabular}
\end{table}

\begin{enumerate}
    \item Range-based Sensitivity

Range-based sensitivity is defined as the difference between the maximum and minimum mAP@0.5:0.95 obtained from the four insertion positions (A1--A4), which measures the variation range of performance:

\begin{equation}
\mathrm{Range} = \max(mAP_{A1:A4}) - \min(mAP_{A1:A4})
\label{eq13}
\end{equation}

\item Standard Deviation (Std)

\begin{equation}
\mathrm{Std} =
\sqrt{
\frac{1}{N}
\sum_{i=1}^{N}
(mAP_i - \overline{mAP})^2
}
\label{eq14}
\end{equation}

where $N = 4$, and $\overline{mAP}$ represents the average mAP across the four insertion positions. Std measures the variation of performance across different insertion positions. Compared with Range, Std is less sensitive to extreme values and provides a more comprehensive measure of model stability of the model under different feature levels.

\end{enumerate}

From Table~\ref{tab3}, it can be observed that GAM exhibits the highest values in both Range and Std, indicating that it is the most sensitive to insertion position. SA shows moderate sensitivity, while Res-CBAM has substantially lower values in both metrics, demonstrating high stability across different feature levels. The consistency between Range and Std further confirms that different attention structures exhibit 
varying degrees of dependence on feature-level variations in the YOLO-AMC architecture, which may influence the final detection performance. In particular, the residual connection in Res-CBAM may provide a stable signal propagation path during feature modulation, thereby reducing the sensitivity of the model to changes in insertion position.

Since cracks usually appear as thin structures with cross-scale distributions, different feature levels preserve different proportions of spatial details and semantic information. Therefore, the feature enhancement effect of the attention module may vary depending on the feature level at which it is applied.

\begin{table}[!ht]
\caption{Comparison of placement sensitivity of different attention mechanisms in terms of Range and Standard Deviation.}
\label{tab3}
\centering
\begin{tabular}{lcc}
\hline
Attention & Range & Std \\
\hline
GAM & 0.0033 & 0.0012 \\
Res-CBAM & 0.0008 & 0.0003 \\
SA & 0.0029 & 0.0010 \\
\hline
\end{tabular}
\end{table}

\subsubsection{Multi-layer Attention Analysis}

In addition to the single-layer insertion configurations (A1–A4), this study further evaluates a multi-layer insertion strategy (A2+A3+A4) to analyze the effect of multi-level feature enhancement in the YOLO-AMC architecture and to investigate whether applying attention modulation at multiple feature levels can produce cumulative improvements. This analysis helps to understand the cooperative effect of attention mechanisms within the multi-level feature pyramid structure. The primary evaluation metric remains mAP@0.5:0.95. As shown in Table~\ref{tab4}, the multi-layer insertion strategy does not consistently outperform the best single-layer configurations. Among the evaluated attention modules, GAM still achieves the highest mAP@0.5:0.95 under the A2+A3+A4 configuration, indicating that applying attention enhancement across multiple feature levels can maintain strong feature representation capability. However, the performance gain compared with the best single-layer insertion setting remains limited, suggesting that excessive attention modulation across multiple feature levels may not always produce cumulative improvements. Similar observations can also be found in Res-CBAM and SA, where the multi-layer configuration provides only marginal differences compared with their corresponding single-layer results.

\begin{table}[!ht]
\caption{Performance comparison of different attention modules under the multi-layer insertion configuration (A2+A3+A4).}
\label{tab4}
\centering
\begin{tabular}{lc}
\hline
Attention & A2+A3+A4 \\
\hline
GAM & 0.9482 \\
Res-CBAM & 0.9429 \\
SA & 0.9378 \\
\hline
\end{tabular}
\end{table}

To quantify the marginal benefit of the multi-layer configuration compared with the best single-layer configuration, the multi-layer gain ($\Delta$ Gain) is defined as:

\begin{equation}
\Delta
=
mAP_{multi-layer}
-
mAP_{best-single}
\label{eq16}
\end{equation}

\noindent where $mAP_{best-single}$ represents the best single-layer result among A1--A4 for the corresponding attention module. $\Delta$ Gain is used to measure the actual performance improvement brought by multi-layer enhancement. To facilitate comparison, Table~\ref{tab5} summarizes the best single-layer result, the multi-layer result, and the corresponding $\Delta$ Gain for each attention module.

\begin{table}[!ht]
\caption{Comparison of multi-layer gain ($\Delta$ Gain) relative to the best single-layer configuration.}
\label{tab5}
\centering
\begin{tabular}{lccc}
\hline
Attention & Best Single & Multi-layer & $\Delta$ Gain \\
\hline
GAM & 0.9441 & 0.9482 & +0.0041 \\
Res-CBAM & 0.9420 & 0.9429 & +0.0009 \\
SA & 0.9418 & 0.9378 & -0.0040 \\
\hline
\end{tabular}
\end{table}

For GAM, the multi-layer configuration (0.9482) is higher than its best single-layer position A3 (0.9441), indicating that under the global attention modeling mechanism, feature enhancement at multiple semantic levels may provide complementary benefits. The $\Delta$ Gain of +0.0041 shows that the multi-layer configuration can further improve detection performance for the GAM structure. This observation may partly explain the performance improvement achieved by the YOLO-AMC architecture when integrating global attention mechanisms.

In contrast, the performance of Res-CBAM under the multi-layer configuration (0.9429) is only slightly higher than its best single-layer result (0.9420), indicating limited improvement. The $\Delta$ Gain is only +0.0009, suggesting that the marginal benefit of multi-layer enhancement is small. This suggests that the structure already possesses stable feature modulation capability, and the multi-layer configuration does not introduce significant additional gain. Under the YOLO-AMC architecture, this implies that certain attention modules can maintain stable performance even with a single-layer configuration.

For SA, the multi-layer insertion leads to performance decrease (0.9378), which is lower than its best single-layer result at A3 (0.9418). The $\Delta$ Gain is -0.0040, indicating that multi-layer enhancement does not necessarily improve detection performance or feature redundancy. This result suggests that in relatively lightweight attention structures, applying attention at multiple levels simultaneously may introduce feature interference, which localization performance may be adversely affected. This further shows that different attention modules have different adaptability to multi-level feature fusion in the YOLO-AMC architecture.

The results indicate that the effectiveness of multi-layer attention insertion depends on the attention module structure. GAM benefits from multi-layer enhancement, whereas Res-CBAM exhibits stable but limited improvement, and SA may experience performance degradation under multi-layer attention modulation.

\subsubsection{Hybrid Attention Analysis}

Based on the previous analyses of single attention modules and multi-layer insertion configurations, this study further investigates the hybrid integration of different attention modules. By assigning different attention mechanisms to different feature levels, the cooperative effects of multiple attention structures within the feature pyramid is analyzed. This experiment is not intended for final model selection, but to investigate the behavioral characteristics of different attention combinations in crack detection.

According to the previous results, GAM exhibits better performance under a multi-layer configuration, whereas Res-CBAM demonstrates higher stability under a single-layer insertion. Therefore, two attention modules are assigned to different feature levels to construct hybrid attention configurations, which are subsequently compared with single-attention models and baseline models.

Two hybrid configurations are evaluated in this section:

\begin{itemize}
    \item A1 GAM, A2+A3+A4 Res-CBAM
    \item A1 Res-CBAM, A2+A3+A4 GAM
\end{itemize}

\begin{table*}[!t]
\caption{Performance comparison of different hybrid attention configurations.}
\label{tab6}
\centering
\begin{tabular}{lccccc}
\hline
Model & Parameters & mAP@50 & mAP@50:95 & Precision & Recall \\
\hline
GAM+Res-CBAM & 3,484,873 & 0.9913 & 0.9497 & 0.9849 & 0.9861 \\
Res-CBAM+GAM & 3,109,925 & 0.9941 & 0.9568 & 0.9848 & 0.9897 \\
GAM & 3,021,587 & 0.9917 & 0.9506 & 0.9829 & 0.9866 \\
YOLOv11 & 2,590,035 & 0.9833 & 0.9112 & 0.9717 & 0.9720 \\
YOLOv8 & 3,011,043 & 0.9707 & 0.8921 & 0.9738 & 0.9480 \\
\hline
\end{tabular}
\end{table*}

\begin{table*}[!t]
\caption{Performance and complexity comparison of C2PSA integration in YOLO-AMC.}
\label{tab7}
\centering
\begin{tabular}{lcccccc}
\hline
Model & Parameters & mAP@50 & mAP@50:95 & $\Delta$ mAP@50:95 & Precision & Recall \\
\hline
GAM & 2,898,515 & 0.9917 & 0.9506 & -- & 0.9829 & 0.9866 \\
GAM+C2PSA & 3,304,083 & 0.9927 & 0.9523 & +0.0017 & 0.9831 & 0.9882 \\
Res-CBAM & 3,450,139 & 0.9903 & 0.9430 & -- & 0.9817 & 0.9842 \\
Res-CBAM+C2PSA & 3,855,707 & 0.9905 & 0.9454 & +0.0024 & 0.9833 & 0.9848 \\
SA & 2,217,451 & 0.9893 & 0.9398 & -- & 0.9835 & 0.9810 \\
SA+C2PSA & 2,623,019 & 0.9903 & 0.9428 & +0.0030 & 0.9837 & 0.9844 \\
\hline
\end{tabular}
\end{table*}

From Table~\ref{tab6}, the hybrid attention configuration achieve performance comparable to, or slightly better than, single-attention models across most evaluation metrics. In particular, Res-CBAM at A1 and GAM at A2+A3+A4 results in the highest mAP@0.5 and mAP@0.5:0.95, indicating improved localization accuracy, while also bringing gains in both Precision and Recall compared to the single GAM model. These results suggest that the complementary behavior of the two attention modules, where Res-CBAM contributes to stable feature refinement high-resolution feature representations and GAM strengthens global feature modeling during multi-level fusion. By integrating different attention mechanisms across feature levels, the model is able to balance local refinement and global context more effectively.

Nevertheless, although this hybrid design achieves the best overall performance, it is regarded as a structural exploration rather than the primary model in this study. The single GAM model is therefore adopted as the reference architecture due to its simpler design and clearer interpretability, while the hybrid configuration serves to demonstrate the potential benefits of combining multiple attention mechanisms.

\subsubsection{C2PSA Integration Analysis}

In the previous ablation experiments, the original C2PSA module in YOLOv11 was removed to isolate the effect of external attention modules. In this section, C2PSA is reintroduced into the YOLO-AMC architecture to examine its impact on different attention models and to analyze its interaction with external attention mechanisms. As shown in Table~\ref{tab7}, all three attention models exhibit slight improvements in mAP@0.5:0.95 after integrating C2PSA, with gains of +0.0017 for GAM, +0.0024 for Res-CBAM, and +0.0030 for SA. Although these results indicate that C2PSA remains complementary to external attention mechanisms, the overall improvement is relatively limited when such mechanisms are already incorporated. In addition, integrating C2PSA results in a noticeable increase in model parameters, with GAM rising from 2.90M to 3.30M, Res-CBAM from 3.45M to 3.85M, and SA from 2.22M to 2.62M, suggesting that the modest performance gain comes at the cost of increased model complexity.

\subsubsection{Detection Results}

\begin{figure*}[!t]
\centering
\includegraphics[width=\textwidth]{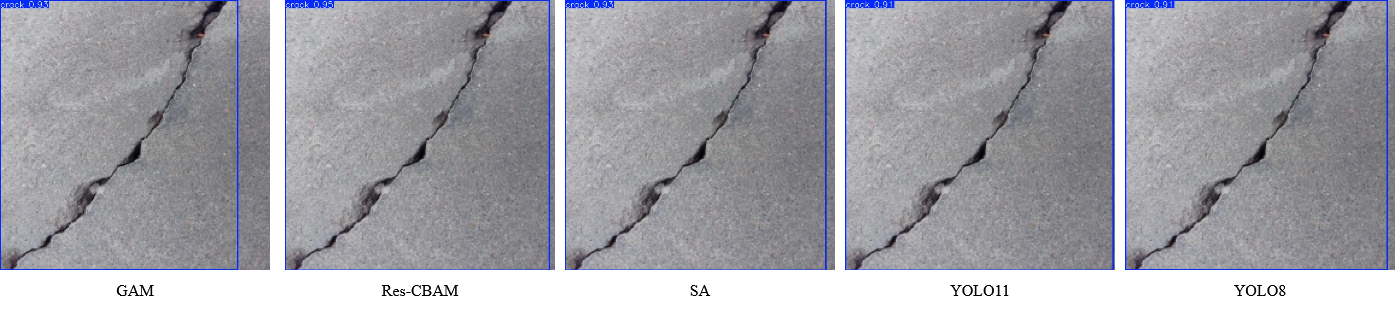}
\caption{Comparison of detection results on a crack image.}
\label{fig3}
\end{figure*}

\begin{figure*}[!t]
\centering
\includegraphics[width=\textwidth]{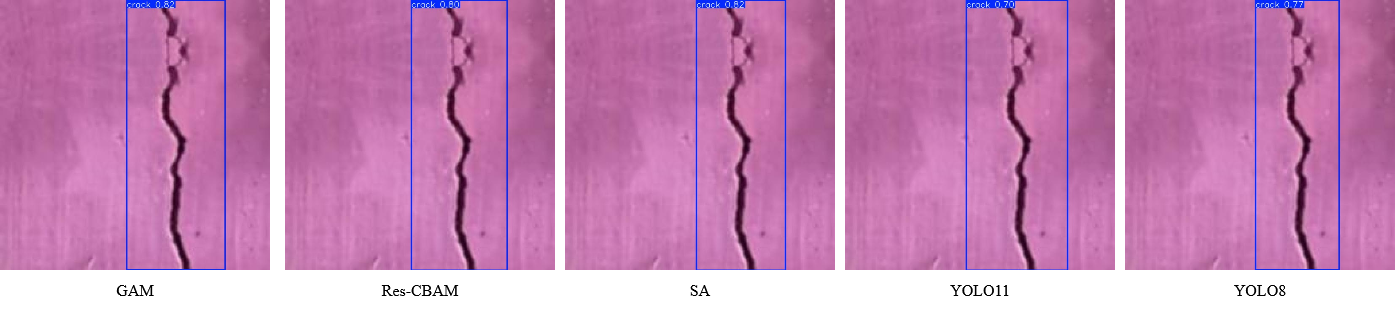}
\caption{Comparison of detection results on a crack image.}
\label{fig4}
\end{figure*}

To provide a more intuitive comparison of the practical performance of different models in crack detection, two representative images from the test set are selected to visualize the detection results of each model, as shown in Figure~\ref{fig3} and Figure~\ref{fig4}. The results compare three YOLO-AMC models (GAM, Res-CBAM, and SA) with two baseline models (YOLOv11 and YOLOv8) on the same input images.

In Figure~\ref{fig3}, the image shows a crack on a normal concrete surface. According to the detection results, all models successfully detect the crack region. The bounding box produced by GAM appears slightly more compact around the crack region, whereas those generated by Res-CBAM and SA are comparable to the baseline models in terms of spatial distribution.

In Figure~\ref{fig4}, the image shows a crack on a concrete surface with a colored coating. The YOLO-AMC variants exhibit higher detection confidence than the baseline models, while the bounding box size and localization remain generally comparable across all models.

The detection results are consistent with the quantitative analysis, showing that attention mechanisms improve crack feature representation.

\subsection{Heatmap Visualization Analysis}

\begin{figure*}[!t]
\centering
\includegraphics[width=\textwidth]{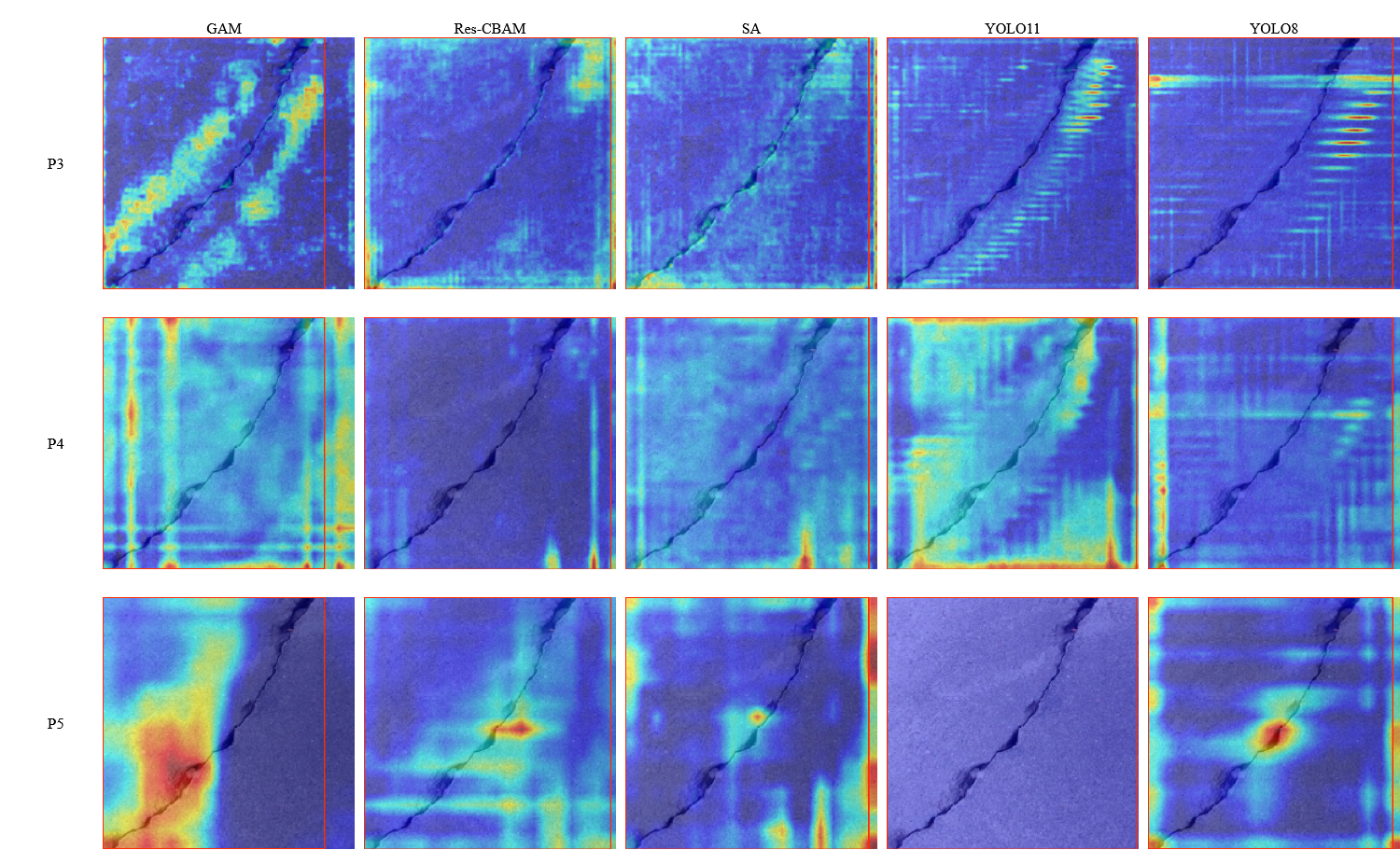}
\caption{Comparison of feature response heatmaps at different feature levels (P3, P4, P5).}
\label{fig5}
\end{figure*}

Grad-CAM++ \cite{chattopadhay2018grad} is employed to visualize the feature responses at different levels of the feature pyramid. To observe the attention distribution in the multi-scale feature pyramid, three feature maps at different semantic representations (P3, P4, and P5), corresponding to low-, mid-, and high-level feature representations, respectively, are selected. The attention-integrated models (GAM, Res-CBAM, and SA) are compared with the baseline models (YOLOv11 and YOLOv8) using the same input image, the resulting feature responses are visualized, as shown in Figure~\ref{fig5}.

\begin{figure*}[!t]
\centering
\includegraphics[width=0.6\textwidth]{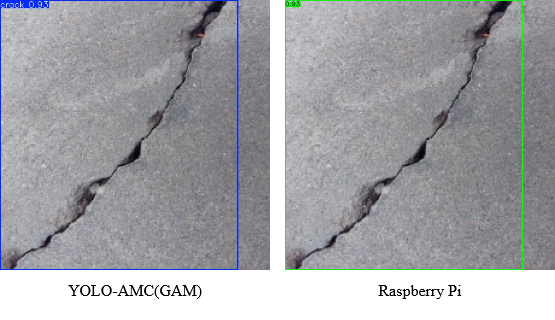}
\caption{Comparison of crack detection results of the YOLO-AMC (GAM) model on GPU and Raspberry Pi 5 platform.}
\label{fig6}
\end{figure*}

At the P3 scale, the heatmaps of YOLOv11 and YOLOv8 exhibit noticeable horizontal and vertical stripe-like patterns, with the horizontal stripes in YOLOv8 being particularly pronounced. These structured response patterns partially obscure the continuity of the crack path. YOLOv11 shows a similar grid-like activation pattern, where high-response regions are distributed along fixed directions rather than being concentrated on the crack geometry. In contrast, the heatmaps of GAM and SA exhibit fewer stripe-like artifacts, and a more continuous response along the crack direction can be observed. However, the activation responses of GAM remain distributed over a relatively broad area and do not show strong local concentration. At this scale, Res-CBAM exhibits a more uniform response pattern, with smaller high-activation regions and lower activation intensity around the crack.

At the P4 scale, the heatmap distributions of different models exhibit more noticeable differences. GAM presents high-contrast activation bands on both sides of the crack, with high-activation regions primarily distributed on the left and right sides of the image, although some background responses can still be observed. The heatmaps of SA and YOLOv11 exhibit stripe-like activations extending along the crack direction, where high-response regions generally align with the crack path. YOLOv8 also shows stripe-like activations near the crack, but activation diffusion can be observed in the background regions. Res-CBAM exhibits relatively low activation intensity at the P4 scale, with large low-response areas in the heatmap and only a few high-activation regions around local crack nodes, suggesting a relatively conservative feature response at this scale.

At the P5 scale, the heatmap response patterns of different models exhibit more pronounced differences. The high-activation regions of GAM are primarily concentrated at the crack intersection in the lower-left portion of the image and extend along the crack direction in a gradient-like manner. The high-activation regions of Res-CBAM converge around local nodes in the middle section of the crack, exhibiting a more localized distribution with relatively low activation in the surrounding background regions. SA exhibits local high-activation regions in the upper-middle section of the crack, and its overall heatmap distribution lies between localized concentration and regional diffusion. YOLOv8 exhibits relatively strong activation responses at the P5 scale, with high-response regions concentrated on local portions of the crack, indicating a tendency toward localized activation. In contrast, YOLOv11 exhibits lower overall activation intensity, with fewer prominent high-activation regions, and its heatmap is characterized by a more uniform low-response distribution.

\subsection{Detection Results on Edge Device}

To verify the practical feasibility of the proposed models on resource-constrained devices, edge-device inference experiments were conducted on a Raspberry Pi 5 platform equipped with a quad-core ARM Cortex-A76 processor (up to 2.4 GHz) and 8 GB RAM, running Raspberry Pi OS. To improve inference efficiency on the edge device, all trained models were converted to the ONNX format. For inference time evaluation, the first 100 images from the validation set were used for benchmarking, and inference was performed in single-image mode with a batch size of 1. To ensure consistency across different platforms, inference time was measured in an end-to-end manner, including image loading, preprocessing, model forward pass, and post-processing operations. The inference performance of each model on the Raspberry Pi 5 is presented in Table~\ref{tab8}.

\begin{table}[!ht]
\caption{Inference performance comparison of different models on Raspberry Pi 5.}
\label{tab8}
\centering
\begin{tabular}{lcc}
\hline
Model & Avg Latency(ms) & FPS \\
\hline
GAM & 202.92 & 4.93 \\
Res-CBAM & 200.31 & 4.99 \\
SA & 178.04 & 5.62 \\
YOLOv11 & 224.77 & 4.45 \\
\hline
\end{tabular}
\end{table}

From the results, all models can successfully run inference on the Raspberry Pi 5 platform, demonstrating that the proposed YOLO-AMC architecture has good feasibility for edge-device deployment. Among them, the SA model achieves the best inference speed, with an average latency of 178.04 ms, corresponding to approximately 5.62 FPS.

To demonstrate the practical detection performance on the edge device, the YOLO-AMC (GAM) model was selected as an inference example, and the results are shown in Figure~\ref{fig6}. The left panel shows the detection result obtained in the GPU environment, while the right panel shows the inference result on the Raspberry Pi 5. The model successfully detects the crack regions on both platforms, and the detection results are highly consistent.

These results demonstrate that the proposed model can be effectively deployed on edge devices while maintaining consistent detection performance.

\section{Discussion}

The experimental results demonstrate that integrating attention modules into the Neck structure can effectively improve the overall performance of crack detection. Compared with YOLOv11 and YOLOv8, the proposed attention-integrated models achieve higher mAP@0.5 and mAP@0.5:0.95. Since cracks are typically thin and low-contrast, their discriminative features is easily weakened during the multi-stage downsampling process. By introducing attention mechanisms in the multi-scale feature fusion stage, the network can re-weight feature responses and focus more on crack-related regions, thereby improving detection performance.

Further observations from Grad-CAM++ \cite{chattopadhay2018grad} heatmaps reveal that different models exhibit distinct response patterns across feature levels. At the P3 scale, baseline models tend to show stripe-like or structured patterns, while attention-integrated models show relatively smoother spatial responses in some regions. At the P4 scale, feature responses gradually align with the crack direction, and attention-based models show more continuous activation patterns in certain cases. At the P5 scale, feature responses become more concentrated, with most models focusing activation on local crack regions, suggesting that high-level features emphasize semantic discrimination.

These phenomena may be related to the characteristics of attention mechanisms and network architecture design. Low-level features are more sensitive to local textures and are more likely to be affected by periodic patterns, which has also been observed in previous studies \cite{zeiler2014visualizing,geirhos2018imagenet}. With the introduction of spatial attention, feature responses can be reweighted across spatial locations, helping to suppress background interference \cite{woo2018cbam}. In addition, mid-level features provide a balance between spatial details and semantic information, making attention mechanisms more effective at this level. The design of different attention modules, such as residual connections and channel attention, may also influence their stability and effectiveness across feature levels \cite{chien2025yolov8am}. At higher feature levels, larger receptive fields tend to focus on more discriminative local regions, which is consistent with findings reported in previous studies \cite{zhou2016learning,selvaraju2017gradcam}.

The effectiveness of attention mechanisms depends not only on whether attention is applied, but also on the module design and its insertion location within the feature pyramid. Different attention structures may exhibit distinct behaviors across semantic levels, indicating that both module characteristics and feature hierarchy should be considered when designing attention-enhanced models, rather than relying on a fixed insertion strategy. In addition, results from the C2PSA integration experiments show that, although C2PSA is complementary to external attention modules, the additional performance gains are relatively limited and accompanied by increased model complexity. These findings suggest that, when external attention mechanisms are already incorporated, the marginal benefits of introducing additional attention modules become less significant.

Overall, attention integration should be regarded as a flexible design strategy rather than a fixed configuration, allowing model architectures to be adapted according to specific application requirements and performance objectives.

\section{Conclusion}

This study proposes an improved YOLO architecture with integrated attention mechanisms, named YOLO-AMC, for building crack detection. Experimental results demonstrate that incorporating different attention modules into the YOLO architecture improves detection performance in terms of mAP@0.5 and mAP@0.5:0.95 compared with YOLOv11 and YOLOv8. These results indicate that attention mechanisms can effectively enhance the representation of fine crack features without introducing significant computational overhead. The insertion location of attention modules within the Neck may also influence detection performance, and the observed variation trends are related to the structural characteristics of the attention modules. However, the impact of insertion location is relatively smaller than that of the attention module design itself.

Based on the above observations, both the selection of attention modules and their insertion locations should be considered when designing attention-enhanced object detection models. Among the evaluated configurations, the GAM-based model demonstrates consistently competitive performance and is therefore selected as the reference model in this study. Meanwhile, hybrid configurations and C2PSA-integrated variants achieve additional performance improvements, indicating that combining multiple attention mechanisms or integrating existing modules can further enhance detection accuracy. The proposed YOLO-AMC exhibits stable detection performance across different crack scenarios and can be successfully deployed on low-power edge devices such as the Raspberry Pi 5, demonstrating the practical feasibility of the proposed approach for real-world applications.

Future work may further investigate the role of attention mechanisms in multi-scale feature fusion and evaluate the generalizability of the proposed design observations across other object detection architectures and application domains.

\bibliographystyle{unsrt}
\bibliography{bibliography}

\end{document}